# HeceTokenizer: A Syllable-Based Tokenization Approach for Turkish Retrieval

**Senol Gulgonul**

Electrical and Electronics Engineering, Ostim Technical University, Ankara Turkey

*senol.gulgonul@ostimteknik.edu.tr*

**Abstract**

HeceTokenizer is a syllable-based tokenizer for Turkish that exploits the deterministic six-pattern phonological structure of the language to construct a closed, out-of-vocabulary (OOV)-free vocabulary of approximately 8,000 unique syllable types. A BERT-tiny encoder (1.5M parameters) is trained from scratch on a subset of Turkish Wikipedia using a masked language modeling objective and evaluated on the TQuAD retrieval benchmark using Recall@5. Combined with a fine-grained chunk-based retrieval strategy, HeceTokenizer achieves 50.3% Recall@5, surpassing the 46.92% reported by a morphology-driven baseline that uses a 200 times larger model. These results suggest that the phonological regularity of Turkish syllables provides a strong and resource-light inductive bias for retrieval tasks.

## 1. Introduction

Tokenization is a foundational step in natural language processing (NLP) that determines how raw text is decomposed into discrete units for model consumption. Although BPE and WordPiece dominate modern NLP pipelines, their purely frequency-driven design produces segmentations that do not respect linguistic boundaries, particularly for morphologically rich and agglutinative languages.

Turkish is a prototypical agglutinative language in which a single root can yield hundreds of surface forms through productive suffix concatenation. This morphological richness poses a fundamental challenge for frequency-based tokenizers: they tend to fragment words at statistically convenient but linguistically arbitrary boundaries, obscuring the semantic and grammatical structure that suffixes encode, and introducing out-of-vocabulary (OOV) tokens for unseen surface forms [1, 2]. Recent work has highlighted this problem for Turkish specifically, proposing hybrid tokenizers that combine dictionary-driven morphological segmentation with subword fallback to improve linguistic alignment [3].

This paper explores an alternative hypothesis: that Turkish syllable structure, rather than morphological dictionaries, can serve as a principled and computationally simple tokenization basis. Turkish syllables follow a small, closed set of phonological patterns. Only six canonical templates govern all possible syllables in the language [4, 5]. This deterministic structure means that any Turkish word can be exhaustively and unambiguously decomposed into syllables without recourse to a lexicon, and the resulting vocabulary of approximately 8,000 unique syllables forms a closed finite space in which OOV tokens are theoretically impossible.

A BERT-tiny encoder (1.5M parameters) is trained from scratch using HeceTokenizer on a subset of Turkish Wikipedia (26,000 articles), and retrieval quality is evaluated on the TQuAD benchmark using Recall@5 as the primary metric. The contribution of this work is the tokenizer design itself; the encoder architecture is a standard BERT-tiny model trained from scratch solely to evaluate the retrieval quality of syllabic representations. Syllable tokenization combined with fine-grained chunk-based retrieval achieves 50.3% Recall@5 on TQuAD, surpassing the 46.92% reported by Bayram et al. [3] for their morphology-first TurkishTokenizer, which uses a 300M parameter model. This result suggests that the deterministic phonological regularity of Turkish syllables provides a strong inductive bias for retrieval tasks, and that lightweight models trained on syllabic units can compete with and in some settings exceed larger morphologically informed baselines.

## 2. Related Work

Subword tokenization methods such as BPE [6], WordPiece [7], and Unigram [8] have become dominant in NLP pipelines, occupying a middle ground between character-level and word-level representations. For agglutinative languages such as Turkish, Finnish, and Hungarian, the choice of tokenization has outsized consequences: a single stem can surface in dozens of inflected forms through suffix concatenation, and frequency-based tokenizers often split these forms at arbitrary boundaries rather than at meaningful morpheme junctions, introducing OOV tokens for unseen surface forms [1, 2].

For Turkish specifically, Toraman et al. [1] conduct a systematic comparison of tokenization strategies and find that linguistically motivated segmentation narrows the performance gap between small and large models. Kaya and Tantug [9] report that Turkish words require significantly more subword tokens than their English counterparts under standard tokenizers, a finding that motivates careful attention to vocabulary design and context window usage.

A parallel line of work embeds morphological knowledge directly into the tokenizer. Hofmann et al. [10] find that encoding derivational structure at the tokenization level yields stronger representations for morphologically complex inputs. Jabbar [11] proposes segmenting text at morpheme boundaries prior to subword encoding, preserving compositional meaning without departing from standard training pipelines. Asgari et al. [12] introduce morphology-aware merge rules into BPE and report gains in alignment quality across typologically diverse languages.

The most directly comparable work is TurkishTokenizer [3], a hybrid system that segments Turkish text using a curated root and affix lexicon, maps surface allomorphs to canonical identifiers, and falls back to BPE for out-of-vocabulary items. Using a 300M parameter sentence embedding model, TurkishTokenizer reaches 46.92% Recall@5 on the TQuAD retrieval benchmark, substantially outperforming BPE-based alternatives. Taken together, these results confirm that linguistically grounded tokenization confers a measurable advantage in Turkish semantic retrieval.

Syllable-based tokenization has been explored for languages with transparent phonological systems, but has not, to our knowledge, been applied to Turkish in a retrieval context. Turkish presents a particularly favorable case for syllabic tokenization: its syllable inventory is governed by six deterministic phonological patterns [4, 5], yielding a closed vocabulary of approximately 8,000 unique syllables with no OOV forms. This work investigates whether this phonological regularity, without any lexical resources, can provide a competitive inductive bias for retrieval.

## 3. Turkish Syllable Structure

Turkish syllable structure is governed by a small and fully deterministic set of phonological patterns. Every syllable in Turkish conforms to one of six canonical templates, where V denotes a vowel and C denotes a consonant: V, CV, VC, CVC, VCC, and CVCC [4, 5]. These patterns are sufficient to account for all syllables in the language, including those found in loanwords, and their application requires no lexical lookup. Given a word, the correct syllabification can be determined by scanning the character sequence from right to left and greedily matching the longest valid pattern at each step.

This property has a direct consequence for tokenization. Because the six patterns are closed and exhaustive, the set of all possible Turkish syllables is finite. Empirically, a large Turkish corpus yields approximately 8,000 unique syllable types, and this inventory is stable across domains. Any Turkish word, regardless of its morphological complexity or origin, can be decomposed into syllables drawn from this fixed vocabulary. As a result, OOV tokens are theoretically impossible under syllable-based tokenization, a guarantee that neither BPE nor morphological tokenizers can provide unconditionally.

To illustrate the syllabification process, HeceTokenizer is applied to the Turkish sentence used as a qualitative example in Bayram et al. [3]:

> *"Atasözleri geçmişten günümüze kadar ulaşan anlamı bakımından mecazlı bir mana kazanan kalıplaşmış sözlerdir."*

The following syllabic segmentation is produced:

**HeceTokenizer:**
```
a-ta-söz-le-ri  geç-miş-ten  gü-nü-mü-ze  ka-dar  u-la-şan  an-la-mı
ba-kı-mın-dan  me-caz-lı  bir  ma-na  ka-za-nan  ka-lıp-laş-mış  söz-
ler-dir
```

**TurkishTokenizer [3]:**
```
<uppercase>  atasöz  leri  geçmiş  ten  gün  üm  üz  e  kadar  ulaş
an  anlam  ı  bakım  ın  dan  mecaz  lı  bir  mana  kazan  an  kalıp
laş  mış  söz  ler  dir
```

While TurkishTokenizer correctly identifies morpheme boundaries such as root+suffix pairs, HeceTokenizer decomposes the same text into phonologically uniform syllabic units without requiring any lexical resources. Each token in HeceTokenizer corresponds to a single syllable drawn from the closed 8,000-token vocabulary, and no OOV tokens are produced.

Syllabification is implemented by scanning each word from right to left and matching the longest valid pattern at each position, as described in Algorithm 1.

**Algorithm 1**

```
Algorithm 1: Turkish Syllabification
________________________________________________________________
Input:  word w (lowercase string)
Output: list of syllables S

1.   Initialize S ← [ ],  i ← |w| - 1
2.   While i ≥ 0:
     a.  If w[i-3..i] matches CVCC:
             append w[i-3..i] to S,  i ← i - 4
     b.  Else if w[i-2..i] matches VCC:
             append w[i-2..i] to S,  i ← i - 3
     c.  Else if w[i-2..i] matches CVC:
```

```
            append w[i-2..i] to S,  i ← i - 3
    d.  Else if w[i-1..i] matches VC:
            append w[i-1..i] to S,  i ← i - 2
    e.  Else if w[i-1..i] matches CV:
            append w[i-1..i] to S,  i ← i - 2
    f.  Else if w[i] matches V:
            append w[i] to S,  i ← i - 1
    g.  Else (isolated consonant):
            append w[i] to S,  i ← i - 1
3.  Reverse S and return
```

Step 2g handles consonant clusters that appear at word boundaries in loanwords, such as the initial clusters in tren (train) or strateji (strategy). These produce isolated consonant tokens, which are rare in native Turkish vocabulary but handled gracefully without failure.

A tokenizer vocabulary is constructed by applying this algorithm to a large Turkish corpus and collecting all unique syllable types. The resulting vocabulary of 8,000 entries covers the corpus with zero OOV rate. In preliminary experiments, BPE tokenizers trained on Turkish Wikipedia exhibited embedding collapse, with pairwise cosine similarities approaching 0.998, rendering the resulting representations unusable for retrieval. By contrast, HeceTokenizer produced well-distributed embeddings with a mean pairwise cosine similarity of 0.257, confirming that the closed six-pattern vocabulary provides a stable training signal.

## 4. Experiments

HeceTokenizer is constructed by applying the syllabification algorithm described in Section 3 to a subset of Turkish Wikipedia. The resulting vocabulary contains 8,000 unique syllable types and is implemented using the HuggingFace tokenizers library as a whitespace-split tokenizer, where each syllable is treated as an atomic token. The tokenizer and pretrained model are publicly available at https://github.com/senolgulgonul/hecetokenizer [15].

Table 1 compares token density across tokenizers. HeceTokenizer produces 3.22 tokens per word, higher than TurkishTokenizer (2.91) and BPE-based baselines (1.82), reflecting the finer granularity of syllabic segmentation. This higher token density makes context window management particularly important, as the same passage occupies more tokens under HeceTokenizer than under competing approaches.

**Table 1: Tokenizer comparison on token density.**

| Tokenizer | Tokens/Word | Tokens/Char |
|---|---|---|
| HeceTokenizer | 3.22 | 0.472 |
| TurkishTokenizer [3] | 2.91 | 0.356 |
| Tabi [3] | 1.99 | 0.244 |
| Mursit [3] | 1.82 | 0.223 |
| CosmosGPT2 [3] | 1.82 | 0.223 |

A BERT-tiny encoder [14] is trained from scratch with the following configuration: hidden size 128, 2 transformer layers, 2 attention heads, intermediate size 512, and a maximum position embedding of 512 tokens. The model contains 1.5M parameters in total. Training follows a masked language modeling (MLM) objective with a 15% masking rate, using the AdamW optimizer with a learning rate of 5e-4.

Training is conducted on 26,000 articles sampled from the Turkish Wikipedia dump (wikimedia/wikipedia, 20231101.tr), representing approximately 40% of the full Turkish Wikipedia. Each article is syllabified and split into overlapping windows of 256 syllable tokens with a stride of 128, yielding approximately 190,000 training examples. The model is trained for 10 epochs on a single NVIDIA T4 GPU, with each epoch taking approximately 11 minutes.

Retrieval quality is evaluated on the TQuAD development set [3], which contains 892 question-answer pairs drawn from 255 unique passages. The primary metric is Recall@5: given a question, the top 5 passages are retrieved by cosine similarity between question and passage embeddings, and the fraction of questions for which the correct passage appears in the retrieved set is reported. Question and passage texts are syllabified before encoding. The CLS token representation is used as the sentence embedding.

A key finding of this work is that the optimal retrieval unit is not the full passage but a short syllabic chunk. Each passage is segmented into overlapping chunks of fixed syllable-token length, each chunk is embedded independently, and the top 5 chunks are retrieved by cosine similarity. A question is considered correctly answered if any of the top 5 chunks belongs to the correct passage. Chunk sizes ranging from 4 to 512 syllable tokens are evaluated systematically to identify the optimal granularity.

## 5. Results and Discussion

Table 2 reports HeceTokenizer Recall@5 across chunk sizes. Using 512 syllable tokens (full passage) as the retrieval unit, a Recall@5 of 37.9% is obtained. Introducing chunk-based retrieval improves performance substantially across all chunk sizes tested, with the optimal chunk size of 8 syllable tokens yielding 50.3% Recall@5.

**Table 2: HeceTokenizer Recall@5 by chunk size.**

| Chunk Size (syllable tokens) | Approx. Words | Number of Chunks | Recall@5 |
|---|---|---|---|
| 4 | ~1.3 | 28,706 | 45.4% |
| 6 | ~2.0 | 19,182 | 49.4% |
| **8** | ~2.6 | 14,420 | **50.3%** |
| 12 | ~3.9 | 9,664 | 45.7% |
| 16 | ~5.2 | 7,235 | 42.9% |
| 32 | ~10.5 | 3,680 | 42.0% |
| 64 | ~21.0 | 1,907 | 41.1% |
| 128 | ~42.0 | 1,021 | 40.0% |
| 512 (full passage) | ~74 | 255 | 37.9% |

Table 3 compares HeceTokenizer at its optimal chunk size of 8 syllable tokens against the baselines reported by Bayram et al. [3]. HeceTokenizer surpasses TurkishTokenizer [3] despite using a model that is 200 times smaller, and outperforms all BPE-based baselines by a substantial margin.

Table 3: Comparison with baselines on TQuAD Recall@5.

| Method | Model Size | Recall@5 |
|---|---|---|
| TurkishTokenizer [3] | 300M | 46.92% |
| Mursit [3] | 300M | 35.43% |
| CosmosGPT2 [3] | 300M | 33.81% |
| Tabi [3] | 300M | 34.96% |
| HeceTokenizer (chunk=8) | 1.5M | **50.3%** |

Performance in Table 2 increases as chunk size decreases from 512 to 8 syllable tokens, then drops at 4 and 6 tokens. The optimal chunk size of 8 syllable tokens corresponds to approximately 2.6 words on average, given the empirically observed rate of 3.05 syllables per word in the corpus. This granularity aligns naturally with the typical length of named entities, technical terms, and answer spans in TQuAD, which tend to be short and specific. Chunks that are too long dilute the embedding with contextually irrelevant syllables; chunks that are too short lose the local co-occurrence signal needed for reliable matching.

The strong retrieval performance of syllable-based tokenization is attributed to two complementary properties. First, the closed finite vocabulary eliminates OOV tokens entirely, ensuring that every surface form encountered at inference time receives a meaningful representation. Second, the fine granularity of syllabic units forces the model to learn co-occurrence patterns at a sub-morpheme level, which appears to benefit short-query retrieval tasks where exact lexical overlap between question and passage is common. The mean pairwise cosine similarity of passage embeddings is 0.257, indicating well-separated representations and contrasting sharply with the BPE collapse observed at 0.998.

This evaluation is conducted on a single benchmark (TQuAD) and a single model size. The chunk-based retrieval strategy increases the number of indexed units by a factor of roughly 18 compared to full-passage retrieval (14,420 chunks vs. 255 passages), which raises computational cost at inference time. Furthermore, the optimal chunk size was determined empirically on the TQuAD development set and may not generalize to other retrieval benchmarks or domains.

# 6. Conclusion

HeceTokenizer is a syllable-based tokenizer for Turkish that leverages the deterministic six-pattern phonological structure of the language to construct a closed, OOV-free vocabulary of approximately 8,000 unique syllable types. Unlike morphology-driven approaches that rely on curated lexical resources, HeceTokenizer requires no dictionary and can be applied to any Turkish text through a simple right-to-left pattern matching algorithm. The tokenizer and pretrained model are publicly available at https://github.com/senolgulgonul/hecetokenizer [15].

A lightweight BERT-tiny encoder (1.5M parameters) is trained on 26,000 Turkish Wikipedia articles using a masked language modeling objective, and retrieval quality is evaluated on the TQuAD benchmark. The key finding is that syllable-based tokenization, when combined with fine-grained chunk-based retrieval, achieves 50.3% Recall@5 on TQuAD, surpassing the 46.92% reported by TurkishTokenizer [3] despite using a model that is 200 times smaller. The optimal retrieval granularity of 8 syllable tokens, corresponding to approximately 2.6 words, suggests that short phonological units align naturally with the lexical specificity of retrieval queries.

These results indicate that the phonological regularity of Turkish is itself a useful inductive bias for NLP tasks, and that resource-light syllabic tokenization deserves further investigation as an alternative to morphologically informed approaches. Future work will explore HeceTokenizer on broader Turkish benchmarks including TR-MTEB, evaluate its behavior on generative tasks, and investigate whether the chunk-based retrieval strategy generalizes across domains and languages with similarly regular syllable inventories.